\documentclass[11pt]{article}
\usepackage[final]{acl}
\usepackage{times}
\usepackage{latexsym}
\usepackage{iftex}
\ifXeTeX
  \usepackage{xeCJK}
  \IfFontExistsTF{Hiragino Mincho ProN}{\setCJKmainfont{Hiragino Mincho ProN}}{\setCJKmainfont{ipaexm.ttf}}
  \newcommand{\beginCJK}{}\newcommand{\finishCJK}{}
\else
  \usepackage[T1]{fontenc}
  \usepackage[utf8]{inputenc}
  \usepackage{CJKutf8}  % pdfLaTeX (arXiv): one document-wide CJK environment, also covers Japanese bib entries
  \newcommand{\beginCJK}{\begin{CJK}{UTF8}{min}}\newcommand{\finishCJK}{\end{CJK}}
\fi
\usepackage{microtype}
\usepackage{inconsolata}
\usepackage{graphicx}
\usepackage{booktabs}
\usepackage{multirow}
\usepackage{xcolor}
\usepackage{amsmath}
\usepackage{amssymb}

\newcommand{\modelname}{XLaw-JP-27B}

\title{\modelname:\\A Lightweight Open Model Specialized for Japanese Law}

\author{Andrew Shin \\
  Keio University \quad X-Law, Inc. \\
  \texttt{andrew@x-law.jp} \\[6pt]
  \normalsize \url{https://x-law.jp/}}

\begin{document}
\beginCJK
\maketitle

\begin{abstract}
We present \modelname{}, an open-weight model specialized for Japanese law that runs on a single 48\,GB GPU under the
Apache~2.0 licence. It adapts Qwen3.8-27B with hybrid retrieval over all current statutes and 31k case summaries, a
light QLoRA fine-tune on its own verified answers to past bar-exam questions (self-distillation), a contract-specific
LoRA adapter, and four-sample voting. We evaluate it on eight Japanese legal tasks---the multiple-choice section of the
bar examination in 2024, 2025 and 2026, statement-level bar questions (JBE-QA), statute questions (LawQA-JP), the
patent attorney exam, and contract review and drafting built from government model contracts---against open models of
12--33B under identical prompts and retrieval. \modelname{} ranks first on six tasks, ties on one and is third on one
(mean rank 1.3 against 2.8--3.1 for the baselines). It is the only system that passes the bar exam in all three years
(129, 120 and 130 of 175, meeting every subject minimum; no test-year material was used in training), while open
Japanese 32--33B models fail at least one year. A controlled study of each component shows that retrieval, voting and
self-distillation help, and that thinking mode, retrieval-grounded synthetic tuning, self-verification,
per-statement decomposition and more self-distillation do not. The largest remaining gaps to frontier models are in
contract review and patent law. The model is available as a free web
service at \url{https://x-law.jp/} with the weights available at \url{https://huggingface.co/X-LawJapan/xlaw-jp-27b} under Apache~2.0.
\end{abstract}

\section{Introduction}

Large language models now pass professional legal examinations in several jurisdictions, but most strong Japanese
results come from closed models or closed retrieval products whose components are not disclosed
\citep{legalscape2025,legalbrain2025}. Law firms, courts, companies and public bodies that cannot send client
documents to an external API need something else: a model that runs on hardware they control, under a licence that
permits commercial use and modification, and that is good at the range of things Japanese legal work involves---not
one exam, but statutes, case law, patents and contracts.

We build such a model and evaluate it broadly. \modelname{} is a 27B-parameter model (Qwen3.8-27B, Apache~2.0)
specialized for Japanese law through retrieval over the current text of every statute and 31k court-decision
summaries, a light fine-tune on its own verified reasoning, a contract adapter, and voting (\S\ref{sec:method}). We
compare it with the strongest open \emph{lightweight} models we could run on the same hardware---Gemma~4 12B,
LLM-jp-4.1 33B and Qwen3-Swallow 32B, the latter two Japanese-specialized---on eight tasks, with identical prompts and
retrieval for every model (\S\ref{sec:overall}). Because one 77-question exam paper moves by up to 16 points between
random seeds of the same system (\S\ref{sec:variance}), we report every seed and paired bootstrap intervals, and we
build the recipe one component at a time on two clean exam years.

\paragraph{Contributions.}
\begin{itemize}
\item \textbf{A lightweight open model for Japanese law.} Among the open models of 12--33B we evaluated, \modelname{} is
best overall across eight tasks (first on six, tied on one, third on one) and the only one to pass the bar exam's
multiple-choice section in 2024, 2025 and 2026 (129, 120, 130; \S\ref{sec:overall}, \S\ref{sec:main}).
\item \textbf{A Japanese legal evaluation suite} covering the bar exam, statement-level and statute questions, the
patent attorney exam (Benrishi-MC) and two new tasks, OI-Contract Review and OI-Contract Draft, built from the Japan
Patent Office's model contracts with checklist grading (\S\ref{sec:setup}, \S\ref{sec:contracts}).
\item \textbf{What works at this scale.} Retrieval contributes most (+17 to +25 points per exam), voting 10--15 and
self-distillation 3--12; on bar statements retrieval already closes $\sim$87\% of the gap to gold articles, so the
remaining errors are reasoning errors (\S\ref{sec:oracle}).
\item \textbf{What does not.} Five negative results, each with a mechanism: thinking mode, retrieval-grounded
synthetic SFT, self-verification, per-statement decomposition and scaled self-distillation (\S\ref{sec:negative}).
\end{itemize}

\section{Related Work}

\paragraph{LLMs on the Japanese bar exam.}
\citet{choi2023jbar} evaluated GPT-3/4 on five years of the 短答式 (30--40\% accuracy).
In our prior work, \citet{shin2026selfverification} fine-tuned GPT-4.1 on the original exam format with self-verification and reached
94.7/175 on 2024 (pass mark 93). Legal retrieval products report 160 on 2024 \citep{legalscape2025} and 169 on 2025
\citep{legalbrain2025}, with undisclosed models and self-reported scoring. We are not aware of a published
open-weight result on the full section, nor of any LLM result on the 2026 exam.

\paragraph{Japanese legal benchmarks.}
JBE-QA \citep{cao2026jbeqa} turns 2015--2024 bar-exam statements into 3,464 true/false items and evaluates 26
models closed-book; the best open model, Swallow-3.3-70B, reaches 0.731 F1. LawQA-JP \citep{uematsu2026lawqa} has
140 four-choice statute questions with gold articles. The COLIEE statute task \citep{coliee2025nowj} uses
Civil Code statements with gold articles; we use its data for evaluation only. LegalRikai \citep{fujita2026legalrikai}
grades contract revision by experts but evaluates closed models only.

\paragraph{Domain adaptation for law.} Legal LLMs are commonly built by continued pre-training or instruction tuning
on legal text. Our negative result with 15k retrieval-grounded synthetic examples (\S\ref{sec:negative}) shows that
such tuning can erode a strong instruction model's reasoning style; we instead keep the base model's own reasoning and
only reinforce its verified answers.

\paragraph{Techniques.}
We build on hybrid lexical--dense retrieval \citep{robertson2009bm25,ruri2024}, self-consistency
\citep{wang2023selfconsistency}, learning from one's own correct rationales \citep{zelikman2022star}, and QLoRA
\citep{dettmers2023qlora}. Japanese open models include LLM-jp \citep{llmjp2024} and Swallow \citep{fujii2024swallow}.

\section{Setup}\label{sec:setup}

\paragraph{Base model.} Qwen3.8-27B \citep{qwen38} (Apache~2.0), run in FP8 with vLLM \citep{kwon2023vllm} on one
48\,GB NVIDIA L40S. Thinking mode is off and we use the model card's non-thinking sampling (temperature 0.7, top-p
0.8, top-k 20, presence penalty 1.5) with fixed seeds; greedy decoding fell into repetition loops on a third of
truncated answers. Exam answers get an 8,192-token budget: at 4,096, a quarter of answers were cut off mid-reasoning
and scored zero.

\paragraph{Bar exam (短答式).} The 2019--2024 papers (憲法, 民法, 刑法; 77 questions, 175 points per year), the 2025 paper and
the 2026 paper, scored with the Ministry of Justice's rules, including partial credit for multi-part answers and the
2026 順不同 questions. A run passes if it reaches the year's pass mark (93 in 2024, 81 in 2025, 99 in 2026) \emph{and} 40\%
in every subject. 2024, 2025 and 2026 are test years: no training data comes from them. On the 2026 exam
human takers averaged 118.1 and successful candidates 126.2.

\paragraph{Other benchmarks.} JBE-QA (3,464 items; we report MCC, which is not inflated by a ``true'' bias, and F1
for comparison with \citealp{cao2026jbeqa}); LawQA-JP (140); the 弁理士 (patent attorney) multiple-choice exam
2024--2025 (120 questions, pass $\approx$65\%); 判例抽出, generating the holdings summary and cited articles of 93
2022--24 decisions (ROUGE-L, set F1); OI-Contract Review and OI-Contract Draft (\S\ref{sec:contracts}); and JCommonsenseQA, JSQuAD and
JEMHopQA to detect regressions in general Japanese.

\begin{table}[t]
\centering\small
\setlength{\tabcolsep}{3pt}
\resizebox{\columnwidth}{!}{%
\begin{tabular}{lrll}
\toprule
Benchmark & Items & Metric & Test material \\
\midrule
司法試験 短答式 2024 & 77 & points/175 & held out \\
司法試験 短答式 2025 & 77 & points/175 & held out \\
司法試験 短答式 2026 & 77 & points/175 & held out \\
JBE-QA & 3,464 & MCC, F1 & 2015--24 \\
LawQA-JP & 140 & accuracy & statutes \\
Benrishi-MC (弁理士) & 120 & accuracy & 2024--25 \\
判例抽出 & 93 & ROUGE-L, F1 & 2022--24 \\
OI-Contract Review & 93 & checklist recall & held-out half \\
OI-Contract Draft & 6 & row score & held-out half \\
COLIEE (Civil Code) & 511 & accuracy, MCC & eval only \\
JCQA / JSQuAD / JEMHopQA & 1,119 / 4,442 / 120 & acc. / EM & general \\
\bottomrule
\end{tabular}}
\caption{Evaluation suite. Training uses only 2019--2023 exam answers produced by the model itself, contract
material from the training half of the OI contracts, and other government model contracts. JBE-QA overlaps the
2019--2023 papers (and contains 2024), so we report it for comparison with the literature, not as a clean test.}
\label{tab:bench}
\end{table}

\paragraph{Variance.}\label{sec:variance} One exam paper is small: across four seeds the same system spans 10--16
points. We therefore report all four seeds, their mean, the vote over them, and 95\% paired bootstrap intervals over
questions (10,000 resamples) for every comparison we draw a conclusion from.

\paragraph{Contamination check.} The 2019--2024 papers and their answer keys have been public for years, the 2026
paper since mid-2026; the base model's pretraining data is not disclosed. If memorisation drove the scores, older
papers should score higher. They do not (Table~\ref{tab:years}): closed-book, the untuned model scores 69--94 on
2019--2023 (mean 83.4) and 88 on 2026; with retrieval 104--131 and 104. Year-to-year differences are of the same size
as seed noise. This does not prove the absence of contamination, but it shows no advantage on the oldest material.

\begin{table}[t]
\centering\small
\setlength{\tabcolsep}{3.2pt}
\resizebox{\columnwidth}{!}{%
\begin{tabular}{lccccccc}
\toprule
& 2019 & 2020 & 2021 & 2022 & 2023 & 2024 & 2026 \\
\midrule
closed-book & 94 & 79 & 89 & 86 & 69 & 79 & 88 \\
+ rag5 & 111 & 104 & 131 & 104 & 111 & 113 & 104 \\
\bottomrule
\end{tabular}}
\caption{Untuned Qwen3.8-27B by exam year (seed 0, /175). No trend towards older, longer-public papers.}
\label{tab:years}
\end{table}

\section{Method}\label{sec:method}

\paragraph{Retrieval (rag5).} The corpus holds the current text of 8,141 laws (392,422 articles, from e-Gov) and
30,884 court-decision summaries. Candidates from BM25 and Ruri-v3-310m dense retrieval \citep{ruri2024} are pooled and
reranked by ruri-v3-reranker-310m, restricted to the laws of the question's subject. On top of this hybrid retriever
(rag4), rag5 (i) looks up every article named in the question directly, (ii) retrieves for each lettered statement
(ア．イ．…) separately as well as for the whole question, and (iii) allows up to 10 articles and 5 case summaries
(5,500 characters). Against a silver set of articles cited in fully correct answers, statute recall rises from 0.395
(rag4) to 0.502 (rag5); on COLIEE statements with gold articles, hybrid retrieval has hit@3 0.890.
Named-article lookup matters because exam statements often cite an article by number (e.g.\ ``民法第94条第2項'');
the per-statement queries matter because the five statements of one question typically concern different articles
(e.g.\ 事務管理: 697, 702 and 703 in the example of Figure~\ref{fig:examples}). For contract review the same retriever
is restricted to 20 contract-relevant laws (Civil Code, the Subcontract Act and its 2026 successor, the Freelance Act,
Antimonopoly Act, consumer-contract, IP and personal-information statutes, conflict of laws and civil procedure).
Table~\ref{tab:rag} compares the two retrievers with everything else fixed.

\begin{table}[t]
\centering\small
\setlength{\tabcolsep}{3.5pt}
\resizebox{\columnwidth}{!}{%
\begin{tabular}{lcc}
\toprule
& rag4 & rag5 \\
\midrule
Statute recall (silver set) & 0.395 & 0.502 \\
Questions with $\ge$1 silver article & 0.844 & 0.921 \\
Statutes in context per question & 3.0 & 4.9 \\
\midrule
2026: mean of 4 runs / worst / vote@4 & 103.8 / 94 / 126 & 108.3 / 104 / 118 \\
2024: mean of 4 runs / worst / vote@4 & 96.8 / 88 / 108 & 111.8 / 103 / 123 \\
Unanimous questions (2026) & 26/77 & 35/77 \\
\bottomrule
\end{tabular}}
\caption{Retriever ablation (untuned model). rag5 raises every single run above the pass mark; its runs agree more,
so voting adds less, and the two years' votes move in opposite directions---a reminder that single-paper comparisons
of one vote are noisy.}
\label{tab:rag}
\end{table}

\paragraph{Self-distillation (E8).} We sample answers from the untuned model on the 2019--2023 papers (with rag5) and
keep 150 whose final answer matches the official key exactly. These are mixed with 91 contract reviews (\S
\ref{sec:contracts}) and used for QLoRA \citep{dettmers2023qlora} (rank 32, 2 epochs, 29 steps of 16 examples,
learning rate $10^{-4}$, max length 8,192), then merged into the base weights. Training never sees 2024, 2025 or 2026.
We call this model E8 in the ablations; the released \modelname{} is E8 plus the contract adapter of
\S\ref{sec:contracts}, which is applied only to contract requests (exam numbers are E8's).
In one rag5 run (seed 0) the untuned model answers 238 of the 383 questions of 2019--2023 exactly right; E8 uses a
random 150 of these answers verbatim (E10 pools further runs: 327 questions with rag5 and 215 closed-book), with the exam prompt and retrieved context they were
produced with, so that training input matches test input exactly. Keeping the model's own long, statement-by-statement
reasoning (median $\approx$2,300 tokens) is what distinguishes this data from the retrieval-grounded SFT set of
\S\ref{sec:negative}, whose short answers taught the model to stop reasoning. The 91 contract reviews were written by
the same model with the official commentary as a private reference; a first version leaked that reference in 229 of 300
answers (「解説にあるように」) and was discarded, the second filters and rewrites such answers.

\paragraph{Voting.} Four samples with different seeds; the most frequent parsed answer wins (ties: lowest seed;
order-free answers vote on the sorted digits).

\section{Japanese legal tasks at a glance}\label{sec:overall}

\begin{table*}[t]
\centering\small
\resizebox{\textwidth}{!}{%
\begin{tabular}{lcccccccc|c}
\toprule
& \multicolumn{3}{c}{Bar exam (points /175)} & Benrishi-MC & LawQA-JP & JBE-QA$^\S$ & \multicolumn{2}{c|}{OI-Contract} & Mean \\
Model (same prompts and rag5 retrieval) & 2024 & 2025 & 2026 & acc. & acc. & MCC & Review & Draft & rank \\
\midrule
\textbf{\modelname} (vote@4; contract adapter) & \textbf{129} & \textbf{120} & \textbf{130} & \textbf{0.517} & \textbf{0.821} & 0.537 & \textbf{0.240} & \textbf{0.829} & \textbf{1.3} \\
\quad single run (mean of 4) & 115.0 & 105.0 & 115.8 & 0.473 & 0.818 & 0.537 & 0.240 & 0.829 & 1.4 \\
Qwen3.8-27B base (vote@4) & 123 & -- & 118 & 0.483 & 0.800 & 0.515 & 0.240 & 0.816 & \\
\midrule
Gemma 4 12B & 90$^\ddagger$ & 87 & 101 & 0.350 & \textbf{0.821} & 0.534 & 0.228 & 0.794 & 2.8 \\
LLM-jp-4.1 33B (thinking) & 100 & 88 & 92$^\ddagger$ & 0.383 & 0.793 & 0.600 & 0.169 & 0.759 & 2.8 \\
Qwen3-Swallow 32B (thinking) & 96 & 85$^\ddagger$ & 82$^\ddagger$ & 0.383 & 0.800 & \textbf{0.660} & 0.173 & 0.692 & 3.1 \\
\midrule
Pass mark & 93 & 81 & 99 & 0.65 & & & & & \\
Claude Opus 5.5 (closed, likely $>100\times$ larger) & 173 & -- & 171 & -- & 0.886$^\ast$ & -- & 0.475 & 0.978 & \\
\bottomrule
\end{tabular}}
\caption{Eight Japanese legal tasks, same prompts and retrieval for every open model; baselines one run each. Mean
rank over the eight task columns among the four systems above the line (ties averaged). $^\ddagger$below the pass mark
or a subject minimum. $^\S$the same 200 JBE-QA items. $^\ast$no retrieval. Differences of a few points on one exam
paper, and the OI-Contract Review gap to Gemma, are within run-to-run noise (\S\ref{sec:variance}).}
\label{tab:overall}
\end{table*}

Table~\ref{tab:overall} summarises the comparison. \modelname{} ranks first on the three bar-exam papers, the patent
attorney exam and both contract tasks, ties Gemma~4 12B on LawQA-JP and is third on JBE-QA, where the two
Japanese-specialized thinking models lead. The ranking does not depend on voting: compared run for run, \modelname{} is
still first on six tasks (mean rank 1.4). It is also the only system that clears every pass mark: Gemma~4 12B misses
2024, LLM-jp-4.1 33B misses 2026 and Qwen3-Swallow 32B misses 2025 and 2026. Relative to its own base model with the
same retrieval and voting, the specialization adds 6--12 exam points, 0.02--0.03 on LawQA-JP, JBE-QA and Benrishi-MC, and 0.013 on
drafting.

The pattern of wins and losses is informative. Thinking models reason at length about a single statement and win on
JBE-QA; on five-statement exam questions the same behaviour exhausts the budget (\S\ref{sec:broad}). Contract review is
the task on which every open model, ours included, is furthest from the frontier (0.17--0.24 against 0.475).

\section{The bar exam in depth}\label{sec:main}

\begin{table}[t]
\centering\small
\setlength{\tabcolsep}{3.5pt}
\resizebox{\columnwidth}{!}{%
\begin{tabular}{lccc}
\toprule
System (one run unless noted) & 2024 & 2025 & 2026 \\
\midrule
Pass mark & 93 & 81 & 99 \\
\midrule
Gemma 4 12B, closed-book & 69 & -- & 76 \\
Gemma 4 12B + rag4 & 83 & -- & 78 \\
Gemma 4 12B + rag5 & 90 & 87 & 101 \\
LLM-jp-4.1 33B (thinking) + rag5 & 100 & 88 & 92 \\
Qwen3-Swallow-32B (thinking) + rag5 & 96 & 85$^\dagger$ & 82 \\
\midrule
Qwen3.8-27B greedy, closed-book & 60 & -- & 65 \\
\quad + recommended sampling, 8k budget & 79 & -- & 88 \\
\quad + rag4 retrieval & 104 & -- & 105 \\
\quad + rag5 retrieval (mean of 4 seeds) & 111.8 & -- & 108.3 \\
\quad\quad vote@4 & 123 & -- & 118 \\
\quad + self-distillation (E8, mean of 4) & 115.0 & 105.0 & 115.8 \\
\quad\quad \textbf{vote@4 (ours)} & \textbf{129} & \textbf{120} & \textbf{130} \\
\midrule
Claude Opus 5.5, (closed, likely $>100\times$ larger) & 173 & -- & 171 \\
\bottomrule
\end{tabular}}
\caption{Bar exam multiple-choice section, points out of 175, official scoring. All Qwen rows share the base model.
Our final system passes all three years including every subject minimum. 2025 was added for the final
system and the baselines only (--: not run). $^\dagger$fails a subject minimum.}
\label{tab:main}
\end{table}

Table~\ref{tab:main} builds the system up one component at a time. Decoding matters first: greedy decoding loops
and runs out of budget, the card's sampling with an 8k budget lifts closed-book scores by 19--23 points. Retrieval
adds the most (+17 to +25 with rag4); rag5 makes single runs pass reliably (worst of four seeds 104 and 103, versus
94 and 88 for rag4). Self-distillation adds a further 3--8 points per run and voting 10--15.

\paragraph{Is the fine-tune real?} Paired by seed, E8 beats the untuned model in 7 of 8 exam runs (one tie). With
voting on both sides the gain is +12 on 2026 (95\% CI $[-1, +26]$, one-sided $p=0.047$) and +6 on 2024 ($[-8,+22]$);
on the 3,460 JBE-QA items it is significant: MCC +0.028 $[+0.012, +0.045]$, $p=0.001$ (Table~\ref{tab:broad}). The
150 training answers come from earlier papers, and nothing else in the training mix concerns the exam, so we attribute
the gain to self-distillation; we test the dose in \S\ref{sec:negative}.

\paragraph{Comparison with prior work.} 129 on 2024 is 34 points above the fine-tuned GPT-4.1 of
\citet{shin2026selfverification} (94.7) and above o3 and Gemini 2.5 Pro as reported by \citet{legalscape2025}
($\approx$115--118, retrieval not disclosed). Closed retrieval products remain far ahead (160--169).

\paragraph{A third test year.} On the 2025 paper (pass mark 81; human candidates averaged 102.1 and successful ones
110.6 according to prep-school reports), the final system scores 120 with vote@4; with the same retrieval, one run
each of Gemma 4 12B, LLM-jp-4.1 33B and Qwen3-Swallow 32B scores 87, 88 and 85 (Swallow misses a subject minimum and
leaves 14\% of answers unparsed); the closed product of \citet{legalbrain2025} reports 169. Single runs score 98--113, and two of
the four miss the constitutional-law minimum (18 and 19 of 50) despite totals far above 81: subject minimums, not totals,
are the binding constraint for single runs, and voting removes the risk (26/50).

\begin{figure}[t]
\centering
\includegraphics[width=\columnwidth]{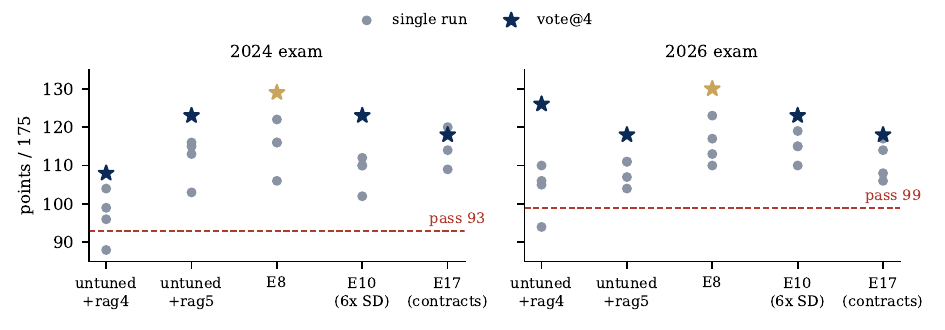}
\caption{Four single runs (dots) and their vote (stars) per system. Seed spread is 10--16 points; voting adds
3--22 points over the mean run, least for systems whose runs already agree (E17).}
\label{fig:seeds}
\end{figure}

\begin{figure}[t]
\centering
\includegraphics[width=\columnwidth]{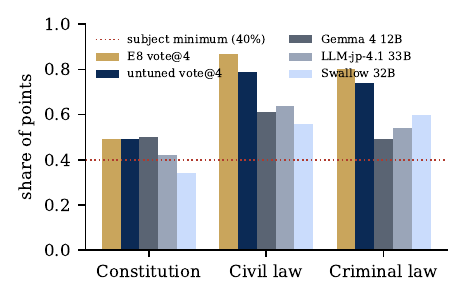}
\caption{Share of points by subject, 2024 and 2026 combined (all systems with rag5; baselines one run).}
\label{fig:subjects}
\end{figure}

\paragraph{Where the points come from.} Pooled over both test years (Figure~\ref{fig:subjects}), the final system
scores 87\% of the civil-law points and 80\% of criminal law but only 49\% of constitutional law---no better than the
untuned model and close to the 40\% subject minimum (22/50 on 2026). Constitutional questions turn on case-law
reasoning about rights and standards of review, which the statute-heavy retrieval supports least; most models we
tested are weakest there. By question format, five-way ``choose the correct pair'' questions are answered best (0.86
of points) and true/false combination tables (one mark per statement) worst (0.33): a table option is right only if all four or five judgements are
right, so one error in five statements costs the whole question. Self-distillation helps most on pairs (+0.08) and
tables (+0.17) and not on per-statement digit answers, where the untuned model is already as good.

\section{Further tasks in detail}\label{sec:broad}

\begin{table*}[t]
\centering\small
\resizebox{\textwidth}{!}{%
\begin{tabular}{lcccccc}
\toprule
& JBE-QA MCC & JBE-QA F1 & LawQA-JP & 弁理士 & 判例 ROUGE-L & JSQuAD EM \\
\midrule
Qwen3.8-27B, closed-book & 0.307 & 0.634 & 0.580$^\dagger$ & 0.300 & -- & 0.875 \\
Qwen3.8-27B + rag5, vote@4 & 0.586 & 0.777 & 0.800 & 0.483 & 0.469$^\ast$ & -- \\
E8 + rag5, vote@4 (ours) & \textbf{0.615} & \textbf{0.785} & \textbf{0.821} & \textbf{0.517} & \textbf{0.514}$^\ast$ & 0.877$^\ast$ \\
E8 + gold articles (oracle) & -- & -- & 0.943 & -- & -- & -- \\
\midrule
Gemma 4 12B + rag5 & 0.598 & 0.786 & \textbf{0.821} & 0.350 & & \\
LLM-jp-4.1 33B (thinking) + rag5 & 0.600$^\S$ & 0.800$^\S$ & 0.793 & 0.383 & & \\
Qwen3-Swallow-32B (thinking) + rag5 & \textbf{0.660}$^\S$ & \textbf{0.835}$^\S$ & 0.800 & 0.383 & & \\
\quad ours on the same 200 items & 0.537$^\S$ & 0.749$^\S$ & & & & \\
\midrule
Best open model in \citet{cao2026jbeqa} (closed-book) & -- & 0.731 & & & & \\
gpt-4.1 in \citet{uematsu2026lawqa} (closed / gold) & & & 0.607 / 0.964 & & & \\
\bottomrule
\end{tabular}}
\caption{Other benchmarks. $^\dagger$mean of 4 seeds; $^\ast$seed 0, no retrieval. 弁理士 is accuracy over 2024--25
(pass $\approx$0.65). JBE-QA closed-book uses all 3,464 items; retrieval rows the 3,460 items in the rag5 cache.
$^\S$the same random 200 JBE-QA items (thinking baselines are 10--20$\times$ slower). Baselines: one run, same
prompts and contexts as ours. Literature rows are for orientation only: their setting differs (no retrieval).}
\label{tab:broad}
\end{table*}

\paragraph{Open baselines.} With the same retrieval, the two Japanese 32--33B thinking models fall short of
the pass mark on 2026 (92 and 82) although they lead on single true/false statements (JBE-QA, +0.06--0.12 MCC on a
shared 200-item sample). Long reasoning helps a single statement but hurts five-statement questions: Swallow
produced no parseable answer for 12\% of exam questions, mirroring our own thinking ablation (\S\ref{sec:negative}).
Gemma 4 12B with rag5 is a strong cheap baseline: one greedy run passes 2026 (101) and ties us on LawQA-JP.

\paragraph{The patent attorney exam.} Benrishi-MC is the hardest benchmark in our suite. With vote@4 the final system
answers 62 of 120 questions (pass mark 78), ahead of the baselines (42--46), but the subject profile is uneven:
copyright and unfair competition 16/20, designs 14/20, trademarks 11/20, treaties 9/20, and patents and utility
models only 12/40---barely above the 20\% expected from guessing among five options. Patent questions hinge on
procedural details (deadlines, divisional and converted applications, priority claims) spread over the Patent Act,
its enforcement orders and the PCT implementation law; we suspect that retrieval reaches the right statute more
often than the right provision (we have no gold articles to check this), and the treaty texts are not in the corpus.

Table~\ref{tab:broad} shows that the fine-tune transfers to statement-level bar questions (JBE-QA) and does not
hurt general Japanese reading comprehension or the judgement summarisation task. It does not close the large gap on
intellectual-property law: the patent attorney exam remains far below its pass mark even with retrieval over the IP
statutes (the corpus lacks the treaty texts that make up a sixth of the exam).

\paragraph{Oracle retrieval.}\label{sec:oracle} On COLIEE Civil Code statements (511, 2019--23; evaluation only),
closed-book MCC is 0.314, rag5 0.598 and gold articles 0.641 (accuracy 0.648 / 0.795 / 0.806): retrieval closes $\sim$87\% of the gap, and
even with the right article the model misjudges about one statement in five. On LawQA-JP's specialized statutes
rag4 retrieval closes only $\sim$53\% of the gap (accuracy 0.580 / 0.770 / 0.936). The bar exam is therefore limited by legal
reasoning, which is why training on the model's own correct reasoning helps there.

\subsection{Contract review and drafting}\label{sec:contracts}

We built two tasks from the twelve OI model contracts of the Japan Patent Office and METI (OIモデル契約書; NDA, PoC,
joint research, licence, AI and data contracts; public data licence). \emph{OI-Contract Review}: given a scenario and one clause presented by
the counterparty, list risks for the client, negotiation points and a revised clause; graded by Claude Opus 5.5
against 3--6 checklist points written from the official commentary. \emph{OI-Contract Draft}: write the whole contract from a
term sheet; each term-sheet row is checked for implementation and consistency. Contracts are split in halves by
document (93 review items, 6 drafts held out).

\emph{Construction.} Each review item is one clause as it appears in the model contract, one of the official
alternative wordings (【オプション】), or a variant proposed in the official commentary (【変更オプション条項】)---so every
clause and every gold point was written by the lawyer-supervised drafting committee, not by us. Claude Opus 5.5 reduced
each clause's commentary to 3--6 atomic, checkable points under three constraints (use only what the commentary says,
no paraphrases of the clause or generic advice, keep the parties' roles straight); items whose commentary yielded fewer
than two reviewable points were dropped, leaving 189 items, and we spot-checked the result. The same model judges
answers point by point. Calibration: Opus itself, answering, covers 0.478 of the points on all items; any system that
reproduced the commentary would score 1.0, so the scale has room above frontier models.

\begin{table}[t]
\centering\small
\begin{tabular}{lcc}
\toprule
Held-out (OI-Contract) & Review & Draft \\
\midrule
Qwen3.8-27B & 0.240 & 0.816 \\
E8 & 0.220 & 0.769 \\
Qwen3.8-27B + statute retrieval & 0.270 & \\
E8 + statute retrieval & 0.241 & \\
Issue-injection SFT (E17) & 0.242 & 0.810 \\
E8 + contract adapter (E17c) & 0.240 & \textbf{0.829} \\
\quad with statute retrieval & 0.247 & \\
Gemma 4 12B & 0.228 & 0.794 \\
LLM-jp-4.1 33B (thinking) & 0.169 & 0.759 \\
Qwen3-Swallow-32B (thinking) & 0.173 & 0.692 \\
\midrule
Claude Opus 5.5 (reference) & 0.475 & 0.978 \\
\bottomrule
\end{tabular}
\caption{Contract review (checklist recall) and drafting (term-sheet row score). Opus also grades, so its own
scores may be inflated by self-preference.}
\label{tab:contracts}
\end{table}

Review is the weakest skill (Table~\ref{tab:contracts}): the model covers about half of what Opus covers and is
weakest on clauses that have been tilted against the client. E8's 91 commentary-grounded reviews did not help.
\emph{Issue injection.} To get training targets whose correctness does not rest on the teacher, we planted one of
26 one-sided problems (uncapped liability for one side, change-of-control termination, assignment of all IP, payment
terms beyond 60 days, unilateral price cuts, \ldots; each with three required review points and the governing
statutes) into 322 sound clauses from government model contracts (MAFF, METI, MLIT) and the training half of the
OI contracts. The model verified each injection (897 of 1,094) and reviewed the altered clause with retrieved statutes
and a private note of the required points; 373 reviews covered every point. Trained from the base model together with
E8's data (E17), review stays at 0.242 and drafting recovers to 0.810, but voting on the exams drops to 118/118
($-12$ and $-11$ against E8, $p\approx0.06$--$0.09$): runs agree more, so the vote adds less. Planted-problem
spotting does not transfer to the commentary-derived checklists, which emphasise startup-specific consequences (M\&A,
listing review, IP strategy). A second, larger round (34 problem types including startup and IP issues; 557 reviews) trained as a
contract-only LoRA adapter on top of E8, so that exams keep using E8 unchanged, improves drafting to 0.829---the best
open result---but leaves review at 0.240 (0.247 with statute retrieval). Every open model and recipe we tried lands
between 0.17 and 0.27 on review, against 0.475 for Opus: closing this gap likely needs expert-written or
stronger-teacher reviews.

\begin{figure*}[t]
\centering\small
\fbox{\begin{minipage}{0.97\textwidth}
\textbf{(a) Bar exam 2026, 民法 Q30 (事務管理, 2 points; correct answer 3 = イ・エ).} Statements include
「ウ．管理者が本人の意思に反して事務管理をしたときは、本人のために有益な費用を支出したとしても、本人に対し、その償還を請求することができない。」
and 「オ．管理者は、自己の財産に対するのと同一の注意をもって事務管理をしなければならない。」\\[2pt]
\emph{E8 (seed 0), excerpts:} 「ウ…誤り。民法702条3項により…本人が現に利益を受けている限度においてのみ…完全に請求できないわけではな」い;
「オ…誤り。民法697条1項後段により、管理者は善良なる管理者の注意…」; for エ: 「…事務管理が成立すると解されます（最高裁平成22年(オ)第43号判決等）。…那是「自分の税務処理」の問題であり…」 → 解答：3 (correct; vote 4/4).\\[2pt]
\emph{Reading.} The verdicts on ウ and オ are grounded in the retrieved articles (702(3), 697). For エ the
conclusion is right but the model adds a docket number that is not in its context and that we could not verify,
and a Chinese phrase (那是, ``that is'') leaks into the Japanese answer. Correct answers can carry unreliable citations.
\end{minipage}}
\vspace{4pt}
\fbox{\begin{minipage}{0.97\textwidth}
\textbf{(b) OI-Contract Review, trademark licence, Art.~8 (本商標).} The licensee (乙) must ``endeavour'' to comply
with the Trademark Act and needs the licensor's consent before using the mark on similar goods or using similar marks.
Checklist (from the official commentary): (1) misuse by the licensee lets anyone demand cancellation of the
registration under 商標法53条; (2) hence the consent requirement in paragraph 2; (3) the ban on conduct that damages
the brand's goodwill (paragraph 3).\\[2pt]
\emph{E8 + contract adapter:} lists missing royalties, unclear scope and term, weak quality-control remedies and
an unclear consent process; proposes a revised clause. Judge: (1)~\texttimes{} (2)~\texttimes{} (3)~\checkmark.\\[2pt]
\emph{Claude Opus 5.5:} 「使用権者の不正使用による取消審判（商標法53条）のリスク…甲が監督していなかったことが取消の根拠になり得ます」,
then genericide and gaps in paragraph 2. Judge: (1)~\checkmark{} (2)~\checkmark{} (3)~\checkmark.\\[2pt]
\emph{Reading.} Our model produces a fluent, generic licence review but misses the statute-specific risk that is the
point of the clause; this is the typical failure behind the 0.24 vs 0.48 gap.
\end{minipage}}
\vspace{4pt}
\fbox{\begin{minipage}{0.97\textwidth}
\textbf{(c) Bar exam 2026, 民法 Q4 (表見代理; correct answer 5).} \emph{Qwen3-Swallow-32B, thinking:} the reasoning
is in English (``We need to determine which statements are incorrect according to case law\ldots''), revisits the
same statement repeatedly and stops at the 12,288-token limit mid-sentence (``\ldots So the rule does not apply, so A is
not bound. But the statement says''), so no answer is given. All nine of its unanswered 2026 questions end this way, all
in civil law. \emph{E8} answers in $\approx$2,500 tokens, statement by statement, and the vote is correct (5; 3 of 4
runs).
\end{minipage}}
\caption{Qualitative examples (excerpts; full outputs in the release materials).}
\label{fig:examples}
\end{figure*}

\section{What did not help}\label{sec:negative}

\begin{table}[t]
\centering\small
\setlength{\tabcolsep}{2.5pt}
\resizebox{\columnwidth}{!}{%
\begin{tabular}{llcc}
\toprule
Technique & Measured on & without & with \\
\midrule
Thinking mode (16k) & 2026 + rag4, one run & 88 & 58 \\
& JBE-QA + rag4 (400), MCC & 0.606 & 0.710 \\
Retrieval-grounded SFT (15k) & 2026 + rag4, one run & 105 & 70 \\
Self-verification & vote@4, 2026 / 2024 & 126 / 108 & 126 / 102 \\
Statement decomposition & E8, 2026 / 2024 & 130 / 129$^\ast$ & 116 / 120 \\
6$\times$ self-distillation & vote@4, 2026 / 2024 & 130 / 129 & 123 / 123 \\
Contract SFT, full model & vote@4, 2026 / 2024 & 130 / 129 & 118 / 118 \\
\bottomrule
\end{tabular}}
\caption{Negative results. Points /175 unless noted. $^\ast$vote@4; decomposition uses five samples per statement,
comparable compute.}
\label{tab:negative}
\end{table}

\paragraph{Thinking mode.} With a 16k budget, thinking raises JBE-QA+retrieval MCC from 0.606 to 0.710 on a
400-item subset, but loses on exams (2026 + rag4: 58 vs 88): it spends 14--15k tokens per multi-statement question
and 27\% of answers are cut off, at 5--10$\times$ the cost.

\paragraph{Retrieval-grounded SFT.} 15,000 synthetic examples (statute true/false, statute QA, consultation, answer
format) from the same model as teacher made every metric worse (2026 + rag4: 70 vs 105). The short training answers
(median 155 characters) taught the model to stop reasoning: exam answers shrank from $\sim$2,300 to 260--350 tokens.

\paragraph{Citation-checked self-verification.} A second turn shows the e-Gov text of every article the draft cites
and asks for a re-check. Of 1,232 answers, 173 changed: 38 wrong$\to$right, 44 right$\to$wrong; vote@4 is unchanged
on 2026 and 6 points worse on 2024. With mostly valid citations (91--96\%), verification mainly induces
second-guessing, unlike the GPT-4.1 setting of \citet{shin2026selfverification}.

\paragraph{Per-statement decomposition.} Judging each statement alone (five samples, own retrieval) and assembling the
answer in code covers 70--76\% of the points and reproduces every official key from perfect verdicts. It scores like a
single run (E8: 116 and 120) and 3--14 points below whole-question voting at similar compute: the model uses
constraints between statements that decomposition hides.

\paragraph{More self-distillation.} Six times more self-distilled answers (877; closed-book and rag5) did not beat
E8 (vote@4 123/123 vs 130/129), although it gave the best constitutional-law scores. The gain from self-distillation
saturates early at this scale.

\section{Cost}\label{sec:cost}

All runs use one NVIDIA L40S (48\,GB; AWS g6e.xlarge, US\$1.86/hour on demand). An exam answer with rag5 averages
about 2,500 output tokens; one 77-question paper takes $\approx$21 minutes, so the four runs of a vote cost
$\approx$US\$2.6 per paper, or about 3.5 cents per question. The thinking baselines generate 2,200 (LLM-jp) to 6,700
(Swallow) tokens per exam question and leave 12\% of Swallow's answers unparsed at a 12k budget. Fine-tuning E8 takes
about one GPU-hour and the whole study, including every negative result, about US\$300 of GPU time plus
$\approx$US\$70 of judge and reference API calls. A single-GPU deployment serving one model around the clock costs about
US\$1,360 per month on demand.

\section{Release and deployment}\label{sec:release}

The weights of \modelname{} (the merged E8 model and the contract adapter, with checksums of every file, the base-model
revision and the training configurations) are publicly released under Apache~2.0. The same weights serve a free public beta, \emph{X-Law JP Agent}, at \url{https://x-law.jp/}. One vLLM
process on a single L40S hosts the merged model and applies the contract adapter per request; a gateway performs the
retrieval of \S\ref{sec:method} for every query (statutes and case summaries; the contract statute set for contract
requests), streams answers, and offers a thinking-mode toggle. Beyond chat, the interface provides clause-by-clause
contract review with Word and Excel reports, contract drafting from a term sheet, and case and statute search. Uploaded
Word, PDF, PowerPoint and Excel files are converted to text, and scanned PDFs and images are read with Japanese OCR
(Tesseract); uploads are kept in memory only and deleted after two hours. A typical answer starts streaming within about
a second and completes in 40--55 seconds; a review of a short contract takes one to two minutes.

\section{Discussion}

A 27B open model, specialized with retrieval and a few hundred examples of its own verified reasoning, is the
strongest of the lightweight open models we tested across eight Japanese legal tasks and passes the bar exam in every
test year. Three lessons generalise beyond this model. First, measure variance before measuring gains: on a 77-question paper,
seed noise is as large as most of the effects in the literature. Second, at 27B the model's own verified reasoning is
better training data than larger volumes of synthetic data from the same model, and a few hundred examples suffice.
Third, the components that help are cheap: hybrid retrieval and four-sample voting run on the same single GPU.

For practitioners, the result means that a model that fits on one workstation-class GPU and can be run on premises
answers bar-exam questions at the level of successful human candidates (126.2 on 2026), provided it retrieves the
statutes. It does not mean the model is reliable: it still misjudges about one statement in five even with the right
article in context, it can attach unverifiable citations to correct conclusions (Figure~\ref{fig:examples}a), and on
contract review it captures half of what a frontier model does. Review of its output by a lawyer remains necessary.
The gaps that remain---constitutional law, intellectual-property law and statute-specific contract risks---are those
where the relevant knowledge sits in case law and specialized commentary rather than in statute text, which suggests
that the next gains will come from retrieval over case law and expert-written training data rather than from more
synthetic data generated by the model itself.

\section*{Limitations}

Three clean exam papers give wide confidence intervals, and component ablations use only two; the 2026 exam was held before the base model's release, so
contamination cannot be fully excluded, although no 2026 material is in our data. The exam scoring is our
implementation of the official rules. Contract grading relies on an LLM judge that is also the reference system. All
baselines run one seed, and open baselines use our prompts and retrieval, which may not suit them. COLIEE results are
pooled over years and are not comparable to official test sets. The system does not give legal advice.

\section*{Ethics Statement}

Exam questions, answer keys, statutes and government model contracts are public-sector information reused under
their terms; COLIEE data was used for evaluation only, under its agreement. The model can produce confident but wrong
legal statements and should be used with professional oversight.

\bibliography{refs}

\appendix
\section{Prompts and hyper-parameters}\label{app:prompts}
All prompts are in Japanese; the system prompt is 「あなたは日本法の専門家です。」 unless noted. With retrieval, the
context block is prepended to the user prompt:

\begin{quote}\small
【参考資料】（条文・判例要旨を検索により取得。関係しないものが含まれる場合があります）\\
◆\textit{law title}\quad\textit{article text, $\le$600 characters} \quad \ldots \quad ◆判例 \textit{summary, $\le$450 characters}\\
---
\end{quote}

\textbf{Bar exam.} 「次の司法試験短答式問題に解答してください。簡潔に検討した上で、最後の行に「解答：」に続けて、
問題文の指示に従った解答番号を数字のみで出力してください。ア・イ・ウ…の各記述についてそれぞれ番号を選ぶ問題では、
記述の順に番号を連続して記載してください（例：アが1、イが1、ウが2なら「解答：112」）。」 followed by the question.
The answer is the digit string after the last 解答：; anything else scores zero.

\paragraph{JBE-QA.} A system prompt asking for 1 if the statement is correct and 0 otherwise (digits only) and the statement as the user turn. Non-binary outputs score as wrong (MCC and F1 treat them as
0).

\paragraph{OI-Contract Review.} The model acts as counsel for the startup (X社) and must give (1) risks for the
client, (2) negotiation points, (3) a revised clause, given an excerpt of the official scenario and the clause. The
judge (Claude Opus 5.5) receives the clause, the numbered checklist and the answer and returns a JSON list of
covered/not-covered verdicts plus whether a revised clause is given; verdicts are cached. \emph{OI-Contract Draft}
asks for the full contract from the term sheet; the judge checks every term-sheet row for implementation and for
consistency with the term sheet and the official model contract.

\paragraph{Decoding.} vLLM 0.30, FP8 weights, thinking off; temperature 0.7, top-p 0.8, top-k 20, min-p 0, presence
penalty 1.5; seeds 0--3; 8,192 new tokens for exams, 4,096 for contract review, 8,192 for drafting, 1,024 otherwise.

\paragraph{QLoRA (E8, E10, E17).} 4-bit base, LoRA rank 32, $\alpha=64$ on all linear layers of the language model,
8-bit AdamW ($\beta=(0.9,0.95)$), learning rate $10^{-4}$ with 5 warm-up steps and decay to 10\%, effective batch
16 sequences, maximum length 8,192, 2 epochs, seed 0. Adapters are merged into the bf16 base by streaming one tensor
at a time and served in FP8. One L40S (48\,GB); E8 trains in about an hour, E17 in 2.3 hours.

\finishCJK
\end{document}